\documentclass[letterpaper, 10 pt, conference]{ieeeconf}  

\IEEEoverridecommandlockouts                              

\usepackage{times}
\usepackage{epsfig}
\usepackage{graphicx}
\usepackage{amsmath}
\usepackage{amssymb}
\usepackage{algorithm}
\usepackage[noend]{algpseudocode}
\usepackage{lipsum}

\usepackage{amsmath}

\usepackage{color}
\usepackage{xcolor}
\usepackage{comment}
\usepackage{graphicx}
\usepackage{booktabs}
\usepackage[margin=2.5cm]{geometry}
\usepackage{caption}

\title{\LARGE \bf
Don't Get Yourself into Trouble! Risk-aware Decision-Making for Autonomous Vehicles}
\author{Kasra Mokhtari$^{1}$ and Alan R. Wagner$^{2}$
\thanks{$^{1}$Department of Mechanical Engineering,
        The Pennsylvania State University, State College, PA 16802, USA
        {\tt\small kbm5402@psu.edu}}%
\thanks{$^{2}$Department of Aerospace Engineering, The Pennsylvania State University,
        State College, PA 16802, USA
        {\tt\small alan.r.wagner@psu.edu}}%
}

\begin{document}

\maketitle
\thispagestyle{empty}
\pagestyle{empty}

\begin{abstract}
\label{Abstract}
Risk is traditionally described as the expected likelihood of an undesirable outcome, such as collisions for autonomous vehicles. Accurately predicting risk or potentially risky situations is critical for the safe operation of autonomous vehicles. In our previous work, we showed that risk could be characterized by two components: 1) the probability of an undesirable outcome and 2) an estimate of how undesirable the outcome is (loss). This paper is an extension to our previous work. In this paper, using our trained deep reinforcement learning model for navigating around crowds, we developed a risk-based decision-making framework for the autonomous vehicle that integrates the high-level risk-based path planning with the reinforcement learning-based low-level control. We evaluated our method in a high-fidelity simulation such as CARLA. This work can improve safety by allowing an autonomous vehicle to one day avoid and react to risky situations. 

\end{abstract}
 
\section{Introduction}
\label{sec1}
Safety is a cornerstone of both Advanced Driver Assistance Systems (ADAS) and autonomous vehicles (AVs). Ensuring human safety is a critical challenge for the widespread deployment of autonomous vehicles on public roads. Although these systems tend to have a wide array of safety features, we believe that risk-based prediction and avoidance of high-risk events are two elements that seem to be underexamined in the existing literature. 

Risk is traditionally described as the expected likelihood of an undesirable outcome~\cite{berger1985prior}. To ensure safety, risk assessment methods that quantitatively estimates the risk associated with an AV's driving options must be developed. To this end, our previous work showed that risk could be characterized by two components: 1) the probability of an undesirable outcome and 2) an estimate of how undesirable the outcome is (loss)~\cite{Mokhtari2019}. Our approach, therefore, considers not only the probability of an accident but also the cost of that accident. For this reason we have developed a general approach to risk-aware decision-making for autonomous systems. We believe that our approach is novel in that it allows for the inclusion of a wide variety of risk types for different vehicles operating in different situations. 

In prior research, we applied our risk assessment method to different types of the autonomous systems, including an AV operating near a college campus and a pair of UAVs flying from Washington DC to Baltimore, to evaluate the risk associated with different options and to select the minimal risk option. We did not focus on the low-level control of those autonomous systems to navigate the least risky traversal. We use the term traversal to refer to traveling along a path at a particular time of the day and day of the week. 

On the other hand, in~\cite{Kasra2020intelligent}, we explored the problem of autonomous vehicle navigation at unsignalized intersections in a structured pedestrian-rich urban environment constructed in a high-fidelity simulation (CARLA)~\cite{dosovitskiy2017carla}. Using a 3D state-space representation of the environment and our conditional reward function, we trained a Double Deep Q-Network (DDQN) integrated with a Prioritized Experience Replay (PER). The resulting control method was capable of safely navigating crowded unsignalized intersections regardless of the intersection topology. This paper presents an integrated system consisting of our risk-based high-level path planning method~\cite{Mokhtari2019} and our lower-level control capable of moving through crowded intersections~\cite{Kasra2020intelligent}. The combination of these two methods allow us to predict risk or potentially risky situations and avoid collisions.

The remainder of the paper is organized as follows: Section \ref{sec2} reviews the related work regarding risk assessment for autonomous vehicles. Section \ref{sec3} describes our risk assessment method to evaluate an autonomous vehicle's traversal and reviews our trained deep reinforcement learning model for AV's navigation among crowds. Section \ref{sec5} introduces the simulation setup and the experiment on which we examine the performance of our risk assessment method. Section \ref{sec6} provides the experimental results and discussion. Finally, Section \ref{sec7} offers conclusions and directions for future research. 
\section{Related Work}
\label{sec2}

Risk can be intuitively understood as the likelihood and severity of the damage that a vehicle of interest may suffer in the future~\cite{lefevre2014survey}. However, in the literature, there are several definitions for risk. For instance, there is a risk associated with the failure of the sensors and the vehicle's components while driving. A fault tree–based risk analysis method is used to distinguish events that could lead to the autonomous vehicles' sensors failure is investigated in~\cite{bhavsar2017risk}. Moreover, risk can be related to the software failures that are implemented in autonomous vehicles. Thieme et al.,~\cite{thieme2020incorporating} describe a process of including software failures in risk analysis that investigates the impact of the propagated failure modes on external interfaces and incorporating these into the risk analysis.

A great deal of research related to risk and autonomous vehicles focuses on minimizing the risk of a collision~\cite{ahangar2021survey}. An intuitive solution to address safety and account for risk is to define a set of rules which characterize the nominal behavior of a vehicle depending on the context, and any deviation from that nominal behavior is considered a danger. For instance, the Chance-Constrained Partially Observable Markov Decision Process (CC-POMDP) model for AVs in urban driving scenarios assumes risk is defined as the probability of the controllable agent violating safety constraints and is implemented to iteratively generate risk-bounded conditional plans over the receding horizon~\cite{huang2018hybrid}. Worrall et al. present a context-based risk assessment method that first generates context area information using the vehicle trajectory and then identifies the safe operation rules~\cite{worrall2010improving}. However, an established limitation of rule-based systems is their inability to account for uncertainties (both on the data and in the model).

As an alternative, others have developed machine learning-based techniques to address a collision risk problem for autonomous vehicles. Chinea and Parent~\cite{chinea2007risk} attempted to assess the risk of a collision by training a Recursive Neural Network (RNN) from simulated driving data. In this work, risk is quantified in terms of the actions and objects present at road intersections including vehicles, pedestrians, buildings, etc. The reliance on simulated data, however, brings into question if this approach will translate to real-world scenarios. Greytak and Hover develop a motion planning controller that incorporates risk of a collision in the motion planning module~\cite{greytak2009motion}. Strickland, Fainekos, and Amor train a deep predictive model on simulated intersection data~\cite{strickland2018deep}. Potential fields methods that include consideration of the risk of collision and use this information to develop optimal controllers have also been considered~\cite{raksincharoensak2016motion}. 

Yu, Vasudevan, and Johnson-Roberson use a Partially Observable Markov Decision Process (POMDP) and reinforcement learning methods to characterize environments that include occlusions and evaluate their method in terms of collision rate and ride comfort on simulated and real-world data~\cite{yu2019occlusion}. Katrakazas, Quddus, and Chen present a method for assessing the risk of autonomous driving at the operational level~\cite{katrakazas2017new}. Specifically, global information is provided as a service to the vehicle which includes information about other vehicles in the area. This information is then used as an aid to control the vehicle by interpreting a traffic situation as either "dangerous" or "safe", enabling a vehicle to be more cautious in situations that are more prone to collisions. David, Lancz, and Hunyady investigated the use of neural networks for real-time risk estimation and different classification algorithms for traffic situation classification. In this context, the risk associated with rapid maneuvers of the ego vehicle is identified based on the probability of a collision (calculated according to Time to Collision (TTC)) and the severity resulted from that collision~\cite{david2019highway}. 

This aforementioned work, although related, generally attempts to use risk to influence the immediate vehicle reactions to a pending collision (collision avoidance), rather than to influence higher-level planning to avoid risky situations. This paper, on the other hand, offers a general framework that integrates a risk-based high-level path planning with the low-level control of autonomous vehicles to avoid risky situations in the first place and then near-collisions with pedestrians. By using a Bayesian Network and taking into consideration causal relationships and conditional probabilities, a general framework is offered which can also be used to create different types of crash accident models. The existing risk assessment methods also tend to rely on simple instantiations of risk~\cite{leveson2011engineering}. In contrast, our method defines and uses risk by integrating a loss function with the conditional probability of various undesirable events, and then selects courses of action in order to minimize that risk. 

\section{Conceptual Framework}
\label{sec3}

We present a method that allows an autonomous vehicle to select the least risky pathway to goal location and negotiate intersections along the way. In this paper, traversal refers to traveling along a path at a particular time of the day and day of the week. Given a predefined set of possible traversal (e.g., times of the day, days of the week, and paths), the autonomous vehicle is expected to navigate from the start point to the endpoint while avoiding collisions with pedestrians. We hypothesize that computing the risk associated with each traversal will allow the AV to avoid risky situations and thus improve safety. This tool serves as a high-level path planner for autonomous vehicles to minimize risk. After the safest traversal is selected, deep reinforcement learning-based controller is employed for the low-level control of the AV's navigation among crowds.  

\subsection{Using Risk to Evaluate an AV's Options}
\label{sec3.1}

Accurately predicting risk or potentially risky situations is critical for the safe operation of autonomous systems. Risk refers to the expected likelihood of an undesirable outcome, such as a collision~\cite{berger1985prior}. In our previous work, we drew on an existing conceptualization of the risk to evaluate an autonomous vehicle's options (e.g. choice of a traversal)~\cite{mokhtari2020don}. In this context, the risk of choosing a traversal $x$ consists of two components: 1) the probability of an undesirable outcome $y$ computed using a Bayesian Network (BN) and 2) an estimate of the loss $L(x,y)$ associated with an undesirable outcome. 

The path of the vehicle is discretized into $N$ timesteps. Therefore, the risk associated with choosing the traversal $x$ at time step $i$ denoted by $R_i(x)$ is calculated as:
\begin{equation} 
\label{eq1}
R_i(x)= \Sigma_{y}^{}{L_i(x,y)p(y\mid I_i)}, 
R(x)= \Sigma_{i=1}^{N}{R_i(x)}
\end{equation}

\noindent where $L_i(x,y)$ is the loss associated with choosing the traversal $x$ at timestep $i$ when event $y$ occurs, $I_i$ is the input set to the Bayesian network at timestep $i$, and $R(x)$ is the total risk associated with choosing the path $x$, respectively. 

\subsubsection{Bayesian Network}
\label{sec3.1.1}

A Bayesian network was designed to calculate the probability of an autonomous vehicle collision with pedestrians. Ideally, the network would be based on data collected from an AV or, perhaps, from high-fidelity simulations. However, since, to the best of our knowledge, the data necessary to construct an accurate Bayesian network related to AV accidents has not been published, we designed a simple yet reasonable network to test the viability of our risk assessment tool~\cite{mokhtari2020don}. 

\begin{figure*}[t]
  \begin{center}
    \includegraphics[scale=0.50]{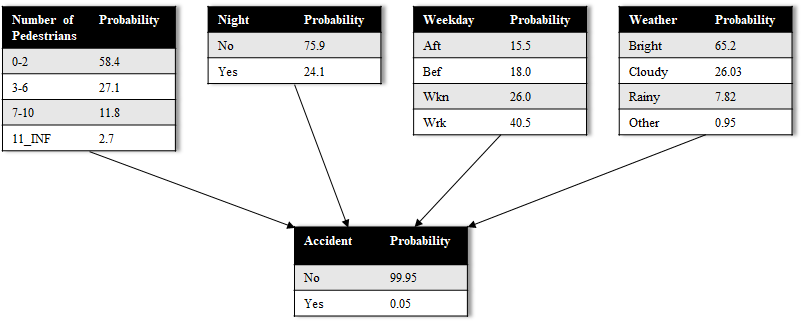}
    \caption{A Bayesian network that attempts to capture the probability of an AV accident. All probabilities were chosen arbitrarily.}
\label{fig23:my_label}
\end{center}
\end{figure*}

The Bayesian network is shown in Fig. \ref{fig23:my_label}. The inputs to the network are the number of pedestrians around the vehicle, the day and time of driving, and the weather. According to Fig. \ref{fig23:my_label}, $Night$ is a Boolean variable. However, $Number of pedestrians$, $Weekday$, and $Weather$ were arbitrarily categorized into four different states. The unconditional probabilities and conditional probabilities were chosen arbitrarily due to the lack of actual data related to autonomous vehicle accidents. 

\subsubsection{Loss Function}
\label{sec3.1.2}

In~\cite{mokhtari2020don}, we defined the loss associated with choosing the traversal $x$ at time step $i$ when the collision (event $y$) occurs as:
\begin{equation} 
\label{eq2}
\begin{split}
L_i(x,y) = Q_i\times W_1  
\end{split}
\end{equation} 
where $Q_i$ is the total number of pedestrians around the vehicle at timestep $i$, and $W_1$ is the constant value assigned to a loss of life, \$10,000,000 (based on recommended insurance coverage of company vehicles). We use dollars as the unit of measure for the loss function in order to make it relatable to other kinds of losses such as property losses. This is standard practice in the risk management literature~\cite{cornett2003financial}.

Because action selection is based on a comparison across available options, as long as the same loss values are used throughout, then these assumptions do not impact the vehicle's decision-making. In fact, loss can be individualized to reflect the values held by actors or a different set of laws. For instance, one autonomous vehicle manufacturer may place more value on accidents resulting in the injury of a person over the destruction of physical property.  

By integrating the conditional probabilities (Section~\ref{sec3.1.1}) with the loss function (Section~\ref{sec3.1.2}), risk is calculated as: 
\begin{equation} 
\label{eq3}
R_i(x)= (Q_i\times W_1) \times p(Accident\mid I_i)
\end{equation}
\begin{equation} 
\label{eq4}
R(x)= \Sigma_{i=1}^{N}{R_i(x)}
\end{equation}

\noindent where $N$ is the number of timesteps for traversal $x$, $I_i$ is the set of inputs to the Bayesian network at timestep $i$, $R_i(x)$ is the risk associated with choosing traversal $x$ at timestep $i$ and $R(x)$ is the total amount of risk associated with choosing path $x$. Note that knowledge of the current environment is assumed at the initial time $i_0$ of the evaluation, and that to evaluate $R_i(x)$ in the future requires a conditional probability assessment of the likelihood of $I_i$ taking on one of many values given known current value $I_{i_0}$ for anytime $i$.

\subsection{Using Reinforcement Learning for AV's Navigation Among Crowds}
\label{sec3.2}

\begin{figure}[t]
    \centering
    \includegraphics[scale = 0.3]{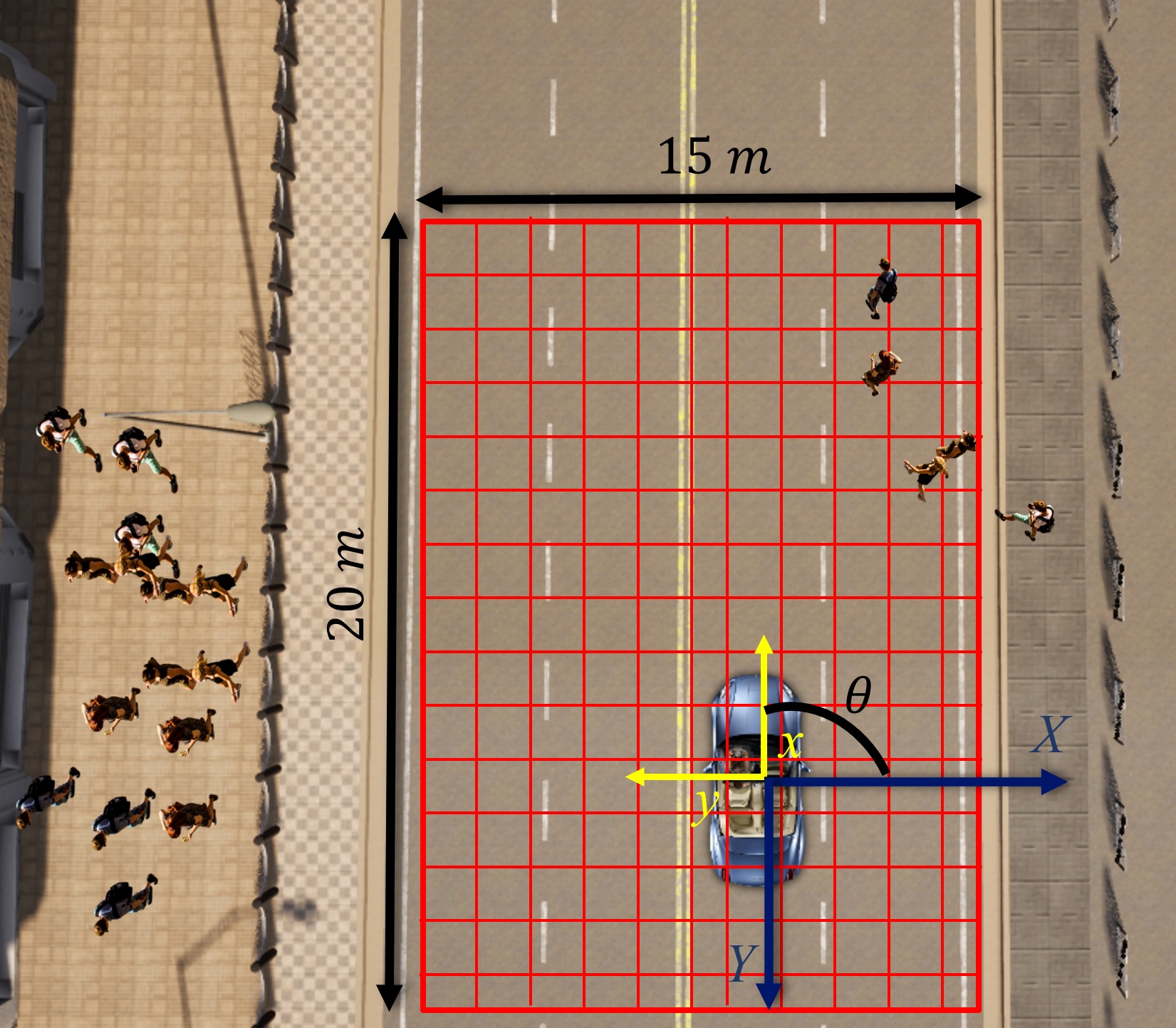}
    \caption{A Bird's eye view of our environment is depicted. The region of interest (ROI) is shown by the red grid (best viewed in color).}
    \label{fig1:intersection}
\end{figure}

In~\cite{Kasra2020intelligent}, we investigated the problem of autonomous vehicle navigation at unsignalized intersections in a structured pedestrian-rich urban environment constructed in CARLA, a high-fidelity autonomous driving simulator~\cite{dosovitskiy2017carla}. The autonomous vehicle's controller was trained on a three-way intersection where the ego vehicle's goal was to make a left-turn without colliding with pedestrians or violating the speed limit. The decision-making process is modeled as an MDP. We utilized a 3D state-space representation and developed a conditional reward function. We then trained a deep reinforcement learning model using Double Deep Q Network (DDQN)~\cite{van2015deep} integrated with Prioritized Experience Replay (PER)~\cite{schaul2015prioritized}. 

The input to the model at timestep $i$ is a multi-layered (3-D) grid. The region of interest (ROI) of the environment around the ego vehicle is a rectangle with length $L$ and width $W$ and is discretized into multiple grids each with the grid discretization $l \times w$ as shown in Figure~\ref{fig1:intersection}. The rectangle is formed such that it always maintains the same orientation with respect to the ego vehicle. In other words, the vertices of the ROI are selected with respect to the ego vehicle's center of gravity which always lies in the cell with an index ($\frac{4L}{5l}$,$\frac{W}{2w}$). We assume that the length and the width of the ego vehicle are $4.5m$ and $2.0m$, respectively. The ROI's parameters are selected as $L=20m$, $W=15m$, $l=0.5m$ and $w=0.5m$. The 3-D tensor state representation contains three 2-D layers where the first 2-D layer represents the cells that are occupied by the ego vehicle and surrounding pedestrians, the relative speed ($m/s$) and the relative heading direction ($degrees$) of the corresponding pedestrians with respect to the ego vehicle are stored in the second and the third tensor layers, respectively. At each timestep $i$, we control the ego vehicle by adjusting the throttle value $a \in [-1,1]$. The action space includes four discrete actions presented in Table~\ref{tab2:action space}. Therefore, the output of the trained deep reinforcement model is one of these four discrete actions. 

Our prior work has empirically demonstrated that our method is not only safe ($100\%$ collision-free episodes) but also capable of successfully ($92-94\%$ completed navigation tasks episodes) navigating at unisignalized intersections in a crowded urban environment regardless of the intersection topology. Our approach can handle different numbers of pedestrians without a noticeable increase in the computation time which makes it suitable for realistic autonomous driving scenarios. Therefore, after identifying the traversal with minimal risk for the autonomous vehicle navigation from the start point to the endpoint (it will be discussed in Section~\ref{sec5}), our trained deep reinforcement model is used to select the low-level control action allowing the AV to navigate among crowds.

\begin{table}[]
\begin{center}
\scalebox{1.2}{
\begin{tabular}{ |c|c|c|c| } 
\hline
Action & Throttle Value & Description \\
\hline
$a_{0}$  & $-1.0$ & $full_brake$\\
$a_{1}$  & $-0.4$ & decelerate\\
$a_{2}$  & $+0.2$ & $accelerate_{1}$\\
$a_{3}$  & $+1.0$ & $accelerate_{2}$\\
\hline
\end{tabular}}
\end{center}
\caption{Action Space}
    \label{tab2:action space}
\end{table}

\section{Experimental Methods}
\label{sec5}

This section uses the risk assessment methods described above to evaluate the autonomous vehicles’ traversals, select a low-risk traversal, and, most importantly avoid high-risk ones. We hypothesized that the methods outlined above would allow the autonomous vehicles to identify traversals that are particularly risky.

\begin{figure}[t]
  \begin{center}
    \includegraphics[scale = 0.35]{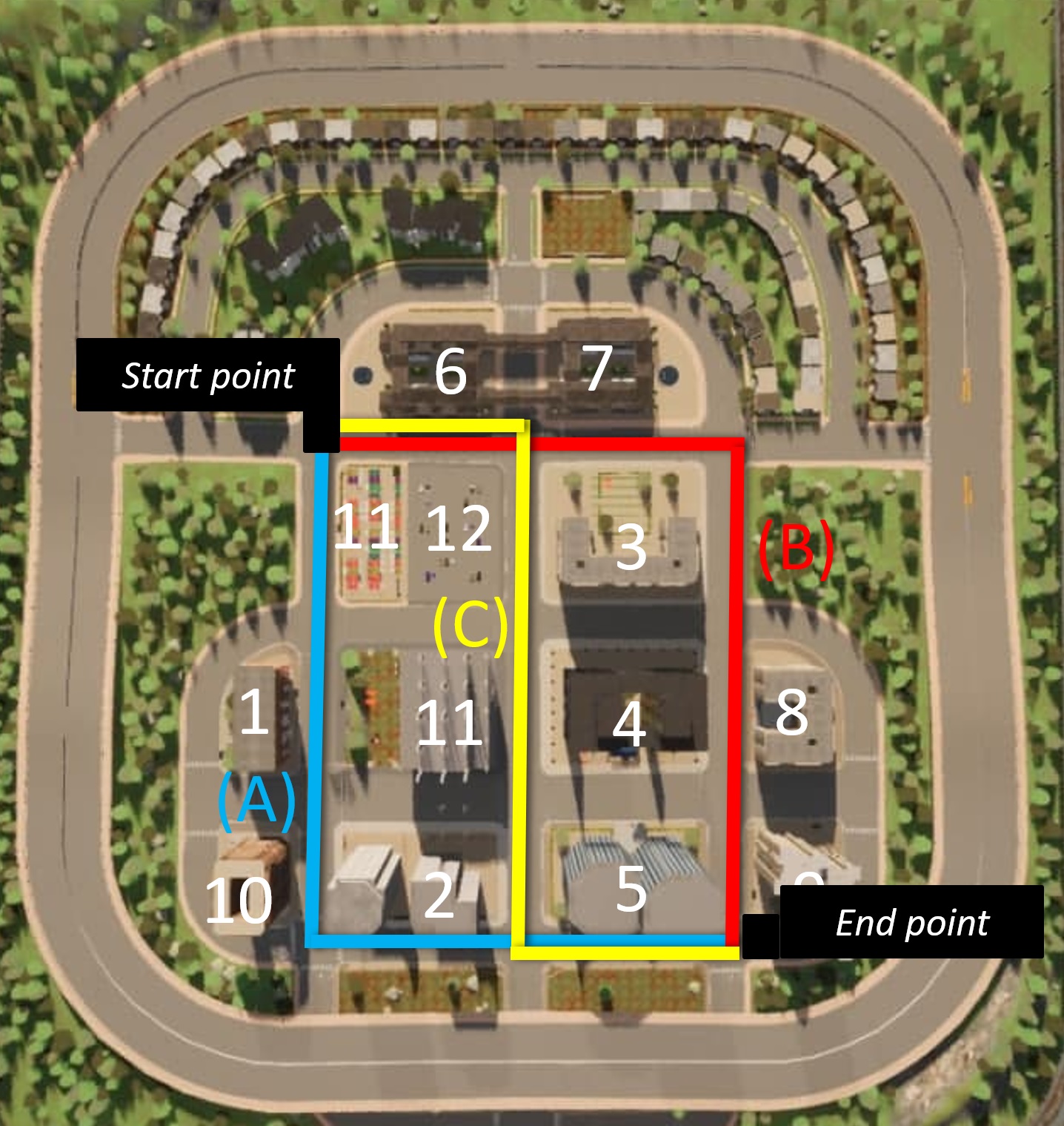}
    \caption{ \small Three different paths between two locations in Town05 in CARLA. The vehicle traversed each path for forty nine times including seven days and seven timeslots a day: 7:00, 9:00, 11:00, 13:00, 15:00, 17:00, and 19:00 (best viewed in color).}
\label{fig12:my_label}
\end{center}
\end{figure}

We used CARLA, an open-source simulator for autonomous driving scenarios, to evaluate the viability of our method. Our environment is depicted in Figure~\ref{fig12:my_label}. Two points were selected as the start point and the endpoint in Town05. Three different paths between these two points were then designed and labeled as Path A, Path B, and Path C as shown in Figure~\ref{fig12:my_label}. The traversals included six times a day (7:00, 9:00, 11:00, 13:00, 15:00, 17:00, and 19:00), seven days of the week (Monday, Tuesday, Wednesday, Thursday, Friday, Saturday, Sunday), and three paths (Path A, Path B, and Path C). As a result, there were one hundred forty-seven traversals between the start point and the endpoint. 

There were different locations along each path that drew crowds depending on the time of the day and day of the week. For instance, location 11 is a bazaar where we assumed that there would be a higher volume of crowds on weekends compared to weekdays. As another example, locations 4 and 5 are shopping malls. Therefore, there were more pedestrians around these buildings during the night relative to the earlier in the morning. Additionally, locations 1, 2, 3, and 8 were commercial properties that saw a surge in pedestrian traffic on weekdays at 9:00 and 17:00 when people normally go to and leave the work. Similarly, locations 11 and 12 are parking lots and they follow the same pedestrian pattern as commercial properties. Moreover, locations 4 and 5 were tall towers. We assumed that the maximum number of people around these two buildings was much higher than that of the small residential buildings such as locations 6, 7, 9, and 10. Depending on the type of the building, a maximum number of pedestrians that could exist around that locations was assigned. The candidate points for spawning the pedestrians around each locations were manually generated. Using the assumptions above, to add randomness to the environment we designed a probability distribution function (PDF) for the number of pedestrians around each location for each combination of time of day and day of the week.

For each traversal, a trajectory was defined for the vehicle by automatically placing a set of waypoints between the start point to the endpoint. In order to compute the risk of selecting a traversal $j$, the traversal was discretized into $N_j$ timesteps where $N_j$ is the length of the traversal in seconds ($s$). For each traversal, using the corresponding PDF and Monte Carl Simulation method, the number of pedestrians around each location was obtained and the pedestrians were then spawned at points that were randomly selected from the set of candidate points around that location. The pedestrians moved to random destination points with a random velocity between 0.2-1.8 (m/s). The $Night$, $Weekday$, and $Weather$ inputs to the Bayesian network were constant at each timestep for each traversal. The input for the $Number of pedestrians$ state varied at each timestep and was then calculated as the number of pedestrians within $20$ meters of the vehicle. We used the same number of pedestrians in equation~\ref{eq2} to compute the loss value for each timestep. Risk for each traversal was then calculated using Equations~\ref{eq3} and \ref{eq4}. We repeated this process one hundred times for each traversal and took the average of these runs to obtain the risk value for the corresponding traversal.  

We consider a situation in which the vehicle can choose either the path, the day of the week, or the time of the day to make the traversal. We, therefore, fixed two of these variables and used the methods described in Section~\ref{sec3} to calculate the risk over the remaining options.

\subsubsection{Which path to travel?}
\label{sec4.1.1}

For this experiment, the autonomous vehicle must select a path (between Path A, Path B, or Path C) for a fixed day and fixed time. Leave-one-out cross-validation was used where the data from a particular day and time for all three paths was left out. We conducted this study by following the steps below: 
\begin{enumerate}
  \item A day and time was selected.
  \item The risk data for the three paths for the selected day and time was removed from the risk data and served as ground truth.
  \item The risk for the three different paths was computed by averaging the risk for all the remaining times and days.
  \item The minimum risk path was chosen as the autonomous vehicle’s best path to travel.
  \item This path was then compared to the data removed in step (2) to determine if the selected path was actually the lowest risk option.  
\item Using the low-level controller discussed in Section~\ref{sec3.2}, the autonomous vehicle navigated along the selected traversal and the number of collisions with pedestrians was recorded. 
\end{enumerate}

This process was repeated 49 times (seven times a day times seven days of the week). The correct number of predictions was divided by 49 to calculate the prediction accuracy.

\subsubsection{Which day to travel?}
\label{sec4.1.2}

A similar procedure was followed to choose the day to travel. In this experiment, the path and the time to travel were fixed and the vehicle selected the day that was the minimum risk traversal. To predict the risk for each day, the risk was averaged with respect to path and time of day. The day predicted to be least risky was chosen. This procedure was followed for all 21 combinations of paths and times of day.   

\subsubsection{Which time to travel?}
\label{sec4.1.3}

Again a similar procedure was followed to determine what time of day to travel. Here, the path and the day to travel were fixed and the vehicle was free to select a time to travel that minimized risk. Seven different times were considered. To predict the risk for each time, the risk was averaged with respect to the path and day of the week. The procedure described above was followed for all 21 combinations of paths and days.

\section{Results and Discussion}
\label{sec6}

\begin{figure*}[t]
  \begin{center}
    \includegraphics[scale=0.65]{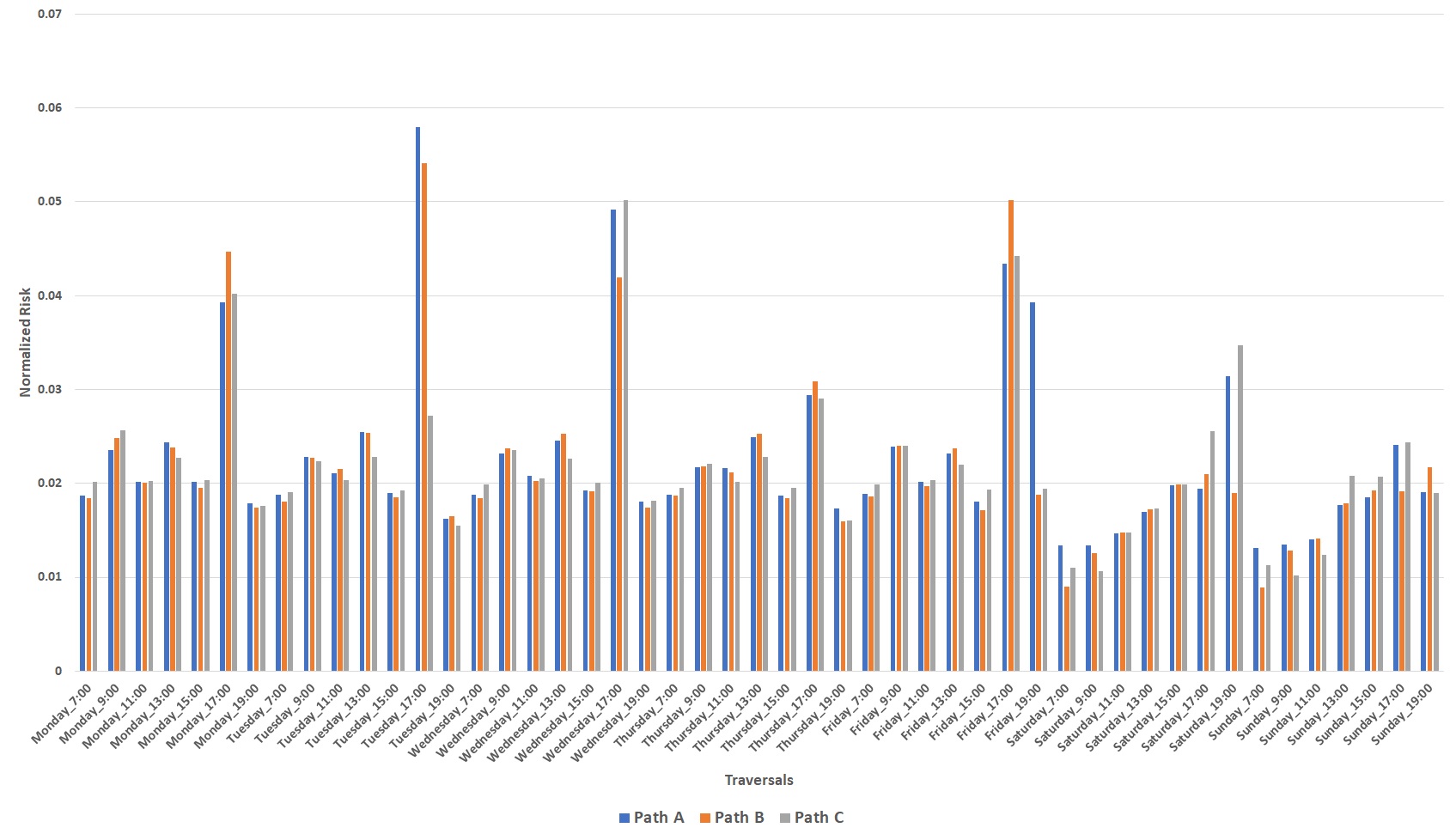}
    \caption{Normalized risk of the all traversals for each path. The spikes imply the important events which the AV might avoid these traversals to increase safety (best viewed in color).}
\label{fig23:my_label}
\end{center}
\end{figure*}

The simulator ran at 15 FPS. Using the trained deep reinforcement learning model as explained in Section~\ref{sec3.2}, a low-level control action for the autonomous vehicle was chosen at each timestep to navigate along each traversal and among the crowds. The risk for a hundred forty-nine traversal was then computed as discussed in Section~\ref{sec5}. \textbf{During the whole experiment, zero collisions with pedestrians were observed}. This confirms that our trained deep reinforcement model was successful at navigating around and through crowds. 

The normalized risk for all the traversals for each path is depicted in Fig~\ref{fig23:my_label}. The risk for each traversal was normalized by dividing by the sum of the risk for all the traversals for the corresponding path. According to Fig~\ref{fig23:my_label}, the majority of which ($\approx 92.0\%$) are below 0.03 in normalized risk. Monday, Tuesday, Wednesday, and Friday at 17:00, and Friday at 19:00 result in a large spike which is two times the average value for Path A. For Path B, the normalized risk on Monday, Tuesday, Wednesday, and Friday at 17:00 are above 0.04. The value for these traversals is approximately two times greater than the average for Path B. For Path C, Wednesday at 17:00 is two times greater than the average for this path. We saw an overall drop in most risk at most times during the weekend. In other words, the risk values for Path B are smaller on weekdays comparing to those on weekends. The risk values for Path A and Path C followed a similar trend except on Saturday 19:00 where a large jump is observed. This occurred because these two paths navigated around locations 4 and 5 which are shopping malls that normally faced a surge in pedestrian traffic on Saturday night. This data clearly demonstrates that specific days and times generate spikes in risk. These spikes are directly related to normal increases in pedestrian traffic during the morning and afternoon on weekdays. More importantly, these spikes represented the risky traversals that an autonomous vehicle should avoid to enhance safety. 

\begin{table}[]
\begin{center}
\scalebox{1.2}{
\begin{tabular}{ |c|c| } 
\hline
Experiment & Prediction Accuracy \\
\hline
Which Path?  & $98.0\%$ \\
Which Day?  & $88\%$ \\
Which Time?  & $72\%$ \\
\hline
\end{tabular}}
\end{center}
\caption{The prediction accuracy for the AV's experiments.}
    \label{tab3:results}
\end{table}

The prediction accuracy for the three experiments described in Section~\ref{sec5} is presented in Table~\ref{tab3:results}. When the autonomous vehicle is tasked with selecting the least risky path, using the method described in Section~\ref{sec5}, the prediction accuracy was $98\%$. This high level of accuracy reflects the fact that one path, normally Path B, tends to avoid pedestrian traffic resulting in lower overall normalized risk. When the vehicle was tasked with choosing the least risky day, the prediction accuracy was $88\%$. Here again, there seems to be a clear risk reduction advantage to driving on weekends.

Finally, when the autonomous vehicle selected the least risky time, the prediction accuracy was $72\%$. In our previous work~\cite{mokhtari2020don}, where we applied the same procedure to our real-world dataset (the Pedestrian Pattern datase~\cite{mokhtari2020pedestrian}), the prediction accuracy was only $19\%$. This previous prediction accuracy reflected the fact that there was not a great risk reduction advantage over the range of times the data was collected at 8:45-17:45. We conjectured that if data had been collected either later in the night or earlier in the morning, the prediction accuracy would have likely increased because early morning hours would have resulted in few pedestrians. In this paper, we conducted our experiment between 7:00 and 19:00. Considering our assumptions for spawning the pedestrians on the map (discussed in Section~\ref{sec5}), the higher prediction accuracy in this paper for selecting the least risky time ($72\%$ compared to $19\%$) confirms our conjecture.


\section{Conclusion}
\label{sec7}

This paper extends our previous work, where we developed a risk assessment tool for computing the anticipated risk over a wide variety of the robot's traversals and selecting the one with the lowest risk. Our previous work, however, did not evaluate the AVs ability to actually travel through each traversal and navigating through and around crowds. Our prior work merely verified that we could pick the least risky path. In contrast, the work here demonstrates that our technique can not only avoid the most risky paths, but it can also handle navigation through crowds if or when it encounters them. This paper thus represents a considerable extension and further development of our previous research.

Our method has shown that some options may offer much greater risk than the average option. To compute risk we have made arbitrary assumptions about the value of loss incurred should an accident occur, the probability distribution function for spawning pedestrians around landmarks, and the Bayesian network. Since action selection is based on comparison across available options, these assumptions do not impact the vehicle's decision-making. Moreover, the risks for this work were computed offline in order for the vehicles to make predictions about the risky events that the vehicle might face in the future, and therefore, it should reconsider its options to enhance safety. Our future work will explore the use of an online risk assessment approach in situations where the risks are rapidly changing and evolving dynamically. We believe that this work can improve safety by allowing an autonomous vehicle to one day avoid and react to risky situations.






{\small
\bibliographystyle{IEEEtran.bst}
\bibliography{bibliography.bib}
}

\end{document}